\documentclass{article}

\usepackage{microtype}
\usepackage{graphicx}
\usepackage{subcaption}
\usepackage{booktabs} 

\usepackage{hyperref}

\usepackage[preprint]{icml2026}

\usepackage{amsmath}
\usepackage{amssymb}
\usepackage{mathtools}
\usepackage{amsthm}

\usepackage[capitalize,noabbrev]{cleveref}

\theoremstyle{plain}

\theoremstyle{definition}

\theoremstyle{remark}

\usepackage[textsize=tiny]{todonotes}

\icmltitlerunning{Patient-Aware Contrastive Learning Preserves Per-Patient Structure in RR-Interval Representations}

\begin{document}

\twocolumn[
  \icmltitle{Patient-Aware Contrastive Learning Preserves Per-Patient Structure in RR-Interval Representations}



  \icmlsetsymbol{equal}{*}

  \begin{icmlauthorlist}
    \icmlauthor{Yasantha Niroshana}{equal,yyy}
    \icmlauthor{Weijith Wimalasiri}{equal,yyy}
    \icmlauthor{Chathuranga Hettiarachchi}{yyy}
  \end{icmlauthorlist}

  \icmlaffiliation{yyy}{Department of Computer Science, University of Moratuwa, Sri Lanka}

  \icmlcorrespondingauthor{Yasantha Niroshana}{yasantha.21@cse.mrt.ac.lk}

  \icmlkeywords{Contrastive Representation Learning, Patient-Aware Representation Learning, Embedding Geometry, Subject-Structured Time Series, Paroxysmal AF Detection, RR Intervals, Wearable Monitoring}

  \vskip 0.3in
]



\printAffiliationsAndNotice{}  
\footnotetext{Code - \href{https://github.com/EML-Labs/pacl-rri-af}{github.com/EML-Labs/pacl-rri-af}}

\begin{abstract}
  Contrastive representation learning struggles on physiological
  signals when each subject contributes a distinct baseline pattern.
  If class differences overlap with subject differences,
  class-level objectives such as supervised contrastive learning tend
  to merge per-subject structure into a single per-class cluster,
  removing the individual variation that a model needs to generalize
  to unseen patients. We study this problem in the setting of
  Paroxysmal Atrial Fibrillation~(PAF) detection from RR-interval~(RRI)
  sequences and propose a \emph{patient-aware contrastive objective}
  that forms positive pairs only from same-patient, same-class
  segments, preserving each patient's own sinus rhythm~(SR) baseline
  while still pushing the two classes apart. Examining the learned
  embeddings directly, our objective achieves the most consistent
  per-patient SR structure (cohesion $0.850$ vs.\ $0.800$ for supervised contrastive loss (SupCon) and $0.772$ for binary cross-entropy (BCE)).
  We also identify that BCE produces the cleanest global class separation
  yet the most disordered per-patient structure. This is precisely
  why a linear probe trained on its features breaks down on unseen
  patients. On the IRIDIA-AF dataset, the resulting representation reaches a patient-independent Area Under the Receiver Operating Characteristic Curve (AUROC) of $0.989 \pm 0.003$ with $2.6\times$ lower seed variance than supervised contrastive baselines.
  These results highlight that per-subject geometric consistency, rather than global class separability, is key to robust cross-patient generalization.
\end{abstract}

\section{Introduction}
Contrastive representation learning is now the default for
unlabeled and weakly-labeled time-series and physiological
signals~\cite{chen_simple_2020,lekhac_contrastive_2020,wang_understanding_2020,khosla_supervised_2021}.
However, its standard formulations assume that every same-class example is equally suitable as a positive. 
This assumption breaks down on data where each example carries both a class label $y$ and a subject identifier $p$. 
When class differences are partly aligned with subject differences, instance-level objectives blend class and identity together.
While class-level objectives such as supervised contrastive learning fold each subject's distinct physiology into a single shared class cluster. 
The resulting embeddings look well-separated overall yet are inconsistent within each subject.
Therefore, a linear probe trained on one cohort fails to transfer to another. 
Our position is that what generalizes across subjects is not how cleanly classes are separated overall, but how consistently each subject is represented.

We study how to build this property directly into the contrastive objective.
We instantiate this question on RRI sequences for
Atrial Fibrillation~(AF) detection: every patient has a
distinct SR baseline, AF dynamics vary across
individuals, and generalizing to unseen patients is the main
bottleneck. Andersen et al.\ reported sensitivity dropping from
$98.98\%$ to $86.04\%$ on held-out patients~\cite{andersen_deep_2019,de_with_temporal_2020,joglar_2023_2024}. 
Self-supervised and contrastive methods have shown promise for Electrocardiogram~(ECG) signals~\cite{grill_bootstrap_2020,liu_dense_2023,hu_novel_2025,sun_enhancing_2025,chen_temporal_2025}, and patient-level objectives have been studied for unsupervised ECG pre-training~\cite{kiyasseh_clocs_2021,diamant_patient_2022}. 
What is missing is a positive-pair construction that is simultaneously class-aware and subject-aware, so it directly targets the per-class, per-subject structure that governs cross-patient transfer.

\textbf{Contributions.}
\begin{itemize}
  \setlength{\itemsep}{0pt}
  \item A \textbf{patient-aware contrastive objective} that, for
    each anchor, treats only same-patient, same-class segments as
    positives, preserving each subject's individual structure
    while still pulling apart classes.
  \item An \textbf{embedding geometry analysis} that explains the
    mechanism: our objective attains the highest per-patient SR
    cohesion among the compared losses. We also identify
    that BCE gives the cleanest global class separation yet the most disordered
    per-patient structure. This is consistent with its weaker transfer to new patients.
  \item \textbf{Downstream validation on AF detection} via
    frozen-encoder linear probing on IRIDIA-AF, reaching AUROC
    $0.989\!\pm\!0.003$ with $2.6\times$ lower seed variance than
    supervised contrastive baselines.
\end{itemize}

\textbf{Related work.}
Contrastive representation learning has progressed from
instance discrimination objectives~\cite{chen_simple_2020,
grill_bootstrap_2020} to label-aware extensions~\cite{khosla_supervised_2021}, with
\citet{wang_understanding_2020} describing the geometry of the
learned representations through alignment and uniformity on the
hypersphere. In the cardiac domain, several patient-level
contrastive variants exist: \textbf{CLOCS}~\cite{kiyasseh_clocs_2021}
contrasts ECG segments across space, time, and patients but is
unsupervised and treats any same-patient pair as a positive;
\textbf{PCLR}~\cite{diamant_patient_2022} uses same-patient
recordings as positives in a SimCLR-style instance task;
\textbf{PMQ}~\cite{sun_enhancing_2025} adds a patient memory queue
to enrich intra-patient comparisons during pre-training. All three
are \emph{unsupervised} pre-training methods that use patient
identity to define positives without using class labels. Our
objective, in contrast, is a \emph{supervised} one whose positive
set is the intersection of class label and patient identity---it
relies on class supervision to prevent supervised contrastive
learning from collapsing each subject into a shared class cluster,
while using subject identity to prevent unsupervised methods from
blending class and identity. To our knowledge, no prior work
applies this combined class-and-subject construction to PAF
detection or analyses the resulting embedding geometry at the
per-subject level.

\section{Methodology}
\label{sec:methodology}

\subsection{Dataset and Preprocessing}

We use IRIDIA-AF~\cite{gilon_iridia_af_2023}, comprising long-term
single-lead ECG recordings from 167 patients with paroxysmal AF.
Episodes are retained when AF duration $\geq 1$~hr and the
immediately preceding SR duration $\geq 4$~hr. Splits are
patient-level (119/24/24 train/val/test); after quality filtering,
154/27/24 episodes remain. RRI sequences are RobustScaler normalized
per-patient using the first SR hour as the fit window, with
classification windows drawn from a strictly disjoint subsequent hour (Appendix~\ref{app:preprocessing} and Figure~\ref{fig:preprocessing} therein). 
The normalized stream is segmented with a sliding window of $W=200$ beats and stride 
$S=50$ beats. Physiologically implausible beats ($<200$\,ms or
$>2000$\,ms) are discarded.

\subsection{Patient-Aware Mini-Batch Sampling}

Each mini-batch samples $P$ patients, each contributing $n$ SR and
$n$ AF windows ($B = 2nP$). A patient is eligible only if both its
SR and AF pools contain at least $n$ windows, guaranteeing
intra-patient, intra-class positive pairs in every batch and
enforcing class balance across patients.

\subsection{A Patient-Aware Contrastive Objective}
\label{subsec:patient_aware_contrastive_objective}
\begin{figure}[t]
  \centering
  \includegraphics[width=\linewidth]{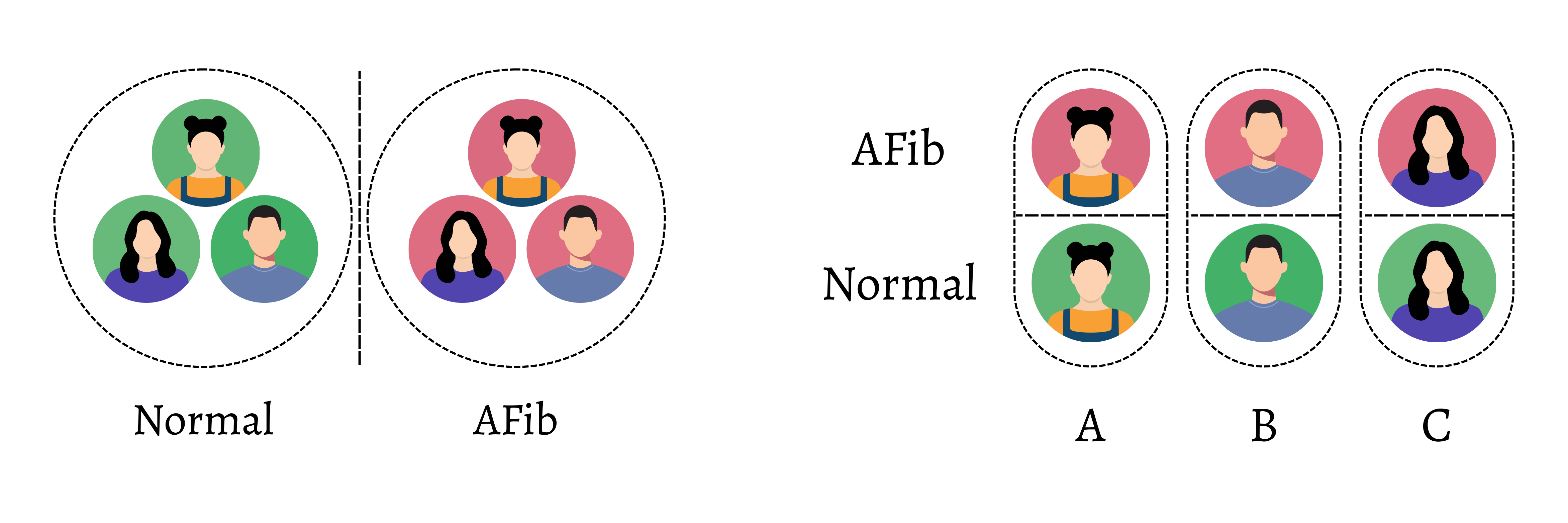}
  \caption{Standard supervised contrastive learning (left) treats all same-class segments as positives regardless of subject. The proposed patient-aware formulation (right) restricts positives to same-patient, same-class segments.}
  \label{fig:framework}
\end{figure}

We formulate the objective as a generic template for
subject-structured data, where each sample carries a class label $y$
and a subject identifier$p$. The positive set is defined to
be \emph{intra-class and intra-subject}. SR/AF (PAF detection) is
the instantiation used here. The construction applies wherever
samples are grouped by subject and the within-subject variation is
informative. For an anchor $i$ with class $y_i$ and subject $p_i$,
the positive $\mathcal{P}(i)$ and negative $\mathcal{N}(i)$ sets are,

\begin{align}
  \mathcal{P}(i) &= \bigl\{\, j \neq i \;\big|\;
    p_j = p_i \;\wedge\; y_j = y_i \,\bigr\}, \\
  \mathcal{N}(i) &= \bigl\{\, j \neq i \;\big|\;
    y_j \neq y_i \,\bigr\}.
\end{align}

Critically, same-class segments from different subjects are
excluded from $\mathcal{P}(i)$, preventing the encoder from
collapsing distinct individual SR baselines into a single shared
prototype. This is the key departure from supervised contrastive
learning~\cite{khosla_supervised_2021}
(Figure~\ref{fig:framework}). The per-anchor loss is the standard InfoNCE formulation,
\begin{equation}
  \mathcal{L}_i = -\log
    \frac{
      \sum_{j \in \mathcal{P}(i)} \exp(s_{ij})
    }{
      \sum_{j \in \mathcal{P}(i)} \exp(s_{ij})
      + \sum_{k \in \mathcal{N}(i)} \exp(s_{ik})
    },
\end{equation}
with $s_{ij} = \hat{z}_i^\top \hat{z}_j / \tau$ and learnable
temperature $\tau$; the batch loss is
$\mathcal{L} = \tfrac{1}{B}\sum_i \mathcal{L}_i$.

\subsection{Encoder and Training}

We use a lightweight multi-branch CNN backbone to demonstrate that the gains come from the loss, not from model capacity. 
This is to remain compatible with edge wearable hardware.
The encoder consists of three parallel 1D-CNN branches
($k\!\in\!\{3,5,7\}$) with stride convolutions (channels 16, 32,
64)~\cite{szegedy_going_2015}, Group Normalization~\cite{wu_group_2018}, and ReLU. 
Branch outputs are fused by a $1{\times}1$ convolution and pooled by a softmax temporal attention module~\cite{shashikumar_detection_2018}. 
A two-layer MLP projects to$\mathbb{R}^{128}$, and both projection and pooled outputs are $\ell_2$-normalized to the unit hypersphere~\cite{wang_understanding_2020}. 
The contrastive loss acts on the projected embedding $\hat{\mathbf{e}}$. The linear probe is trained on the pooled output $\hat{\mathbf{z}}$.
All three loss functions (Proposed, SupCon, BCE) are trained with  the identical patient-aware sampler described in Section~\ref{subsec:patient_aware_contrastive_objective}. The sampler is not varied across conditions.
Training uses AdamW with a cosine scheduler. We use Optuna-TPE hyperparameter optimization~\cite{akiba_optuna_2019} to select the hyperparameters on the validation split (full configuration in Appendix~\ref{app:training}).

\section{Experiments}

\textbf{Protocol.}
We adopt frozen encoder linear probing, a standard
representation quality readout in contrastive
learning~\cite{wang_understanding_2020,chen_simple_2020,khosla_supervised_2021}. 
Probe performance reflects the embedding geometry rather than classifier capacity. 
All experiments are repeated over five independent seeds and reported as mean~$\pm$~std on the held-out patient-independent test split.

\subsection{Embedding Geometry}
\label{subsec:repr_analysis}

We probe the embedding geometry directly. We compare the proposed
objective against SupCon~\cite{khosla_supervised_2021} and BCE baselines, fixing the encoder, sampler, and
probe. Only the temperature (when applicable) $\tau$ and learning rate is varied to find the best performance.
We measure,
\begin{itemize}
  \setlength{\itemsep}{0pt}
  \item Per-patient \emph{class cohesion}, the mean cosine similarity of same-patient same-class embeddings.
  \item Global class separability via centroid distance, centroid cosine similarity, and compactness ratio 
  \item Per-patient compactness ratio. 
\end{itemize}
Full definitions are in Appendix~\ref{app:metric_defs}.
Results are summarized in Table~\ref{tab:repr_metrics}.
SR cohesion is the primary metric because sinus rhythm is each 
patient's individual baseline which is the reference point a cross-patient 
linear probe must extrapolate from when it encounters a new subject.High AF cohesion without corresponding transfer improvement confirms 
that SR placement consistency, not AF compactness, is the bottleneck.

\begin{table}[t]
  \caption{Embedding-space metrics on the test set (mean~$\pm$~std, 5 seeds). Per-patient SR cohesion is the primary metric. It directly measures the per-subject geometric consistency the objective targets. The bottom row reports downstream linear-probe AUROC. Bold = best per row. $\uparrow$ higher is better, $\downarrow$ lower is better.}
  \label{tab:repr_metrics}
  \centering
  \scriptsize
  \setlength{\tabcolsep}{3pt}
  \begin{tabular}{lccc}
    \toprule
    \textbf{Metric}
      & \textbf{Proposed}
      & \textbf{SupCon}
      & \textbf{BCE} \\
    \midrule
    SR cohesion $\uparrow$
      & $\mathbf{0.850\!\pm\!0.044}$
      & $0.800\!\pm\!0.029$
      & $0.772\!\pm\!0.049$ \\
    AF cohesion $\uparrow$
      & $0.846\!\pm\!0.048$
      & $0.921\!\pm\!0.035$
      & $\mathbf{0.955\!\pm\!0.019}$ \\
    Centroid sim.\ $\downarrow$
      & $+0.238\!\pm\!0.200$
      & $+0.155\!\pm\!0.123$
      & $\mathbf{-0.717\!\pm\!0.029}$ \\
    Centroid dist.\ $\uparrow$
      & $1.034\!\pm\!0.100$
      & $1.120\!\pm\!0.095$
      & $\mathbf{1.603\!\pm\!0.045}$ \\
    Global compact.\ $\uparrow$
      & $1.192\!\pm\!0.124$
      & $1.313\!\pm\!0.465$
      & $\mathbf{2.427\!\pm\!0.226}$ \\
    Per-pt.\ compact.\ $\uparrow$
      & $\mathbf{3.273\!\pm\!0.433}$
      & $2.178\!\pm\!1.148$
      & $6.396\!\pm\!0.590$ \\
    \midrule
    \textbf{AUROC} $\uparrow$
      & $\mathbf{0.989\!\pm\!0.003}$
      & $0.983\!\pm\!0.009$
      & $0.980\!\pm\!0.012$ \\
    \bottomrule
  \end{tabular}
\end{table}

\textbf{Per-patient SR cohesion confirms the design.}
The proposed objective achieves the highest SR cohesion
($0.850$ vs.\ $0.800$ SupCon vs.\ $0.772$ BCE), evidence that
intra subject positives preserve each subject's own SR baseline rather
than collapsing them into a shared class prototype. The AF cohesion
remains balanced ($0.846$, vs.\ $0.921$ SupCon, $0.955$ BCE),
indicating that the SR gains do not come at the cost of disordering the AF cluster. 
The objective preserves both classes' per subject structure, in contrast to SupCon's lopsided trade-off.

\textbf{The BCE paradox.}
BCE produces the most clearly separated class means in the
embedding space (centroid distance $1.603$, cosine similarity
$-0.717$) and the highest global compactness ratio ($2.427$). However, it
gives the worst downstream AUROC on unseen patients
(Section~\ref{subsec:downstream_paf}). The reason becomes clear
once we look inside each patient. Those well separated class means
coexist with disorganized per-patient SR placement ($0.772$
cohesion). A linear probe trained on seen patients has no
consistent SR direction to extrapolate from when it meets a new
one. This is consistent with the hypothesis that what governs transfer 
to new subjects is per-subject consistency, not how cleanly the 
classes are separated overall.. BCE achieves the
highest \emph{per-patient} compactness ratio ($6.396$). However, it is only because its denominator, the within-patient spread of each class shrinks toward zero.
The positions of those tight clusters then drift unpredictably from one patient to the next, which is what the linear probe actually has to follow.

\textbf{SupCon over-compacts AF, under-aligns SR.}
SupCon achieves the tightest AF cohesion ($0.921$) by pulling all AF segments together regardless of subject, at the direct cost of SR cohesion ($0.800$ vs.\ $0.850$). 
This is the predicted failure mode. 
Standard supervised contrastive learning trades per-patient SR structure for global AF compactness, precisely the tradeoff that hurts patients with atypical SR baselines.

\subsection{Downstream PAF Detection}
\label{subsec:downstream_paf}

\begin{figure}[t]
  \centering
  \includegraphics[width=\linewidth]{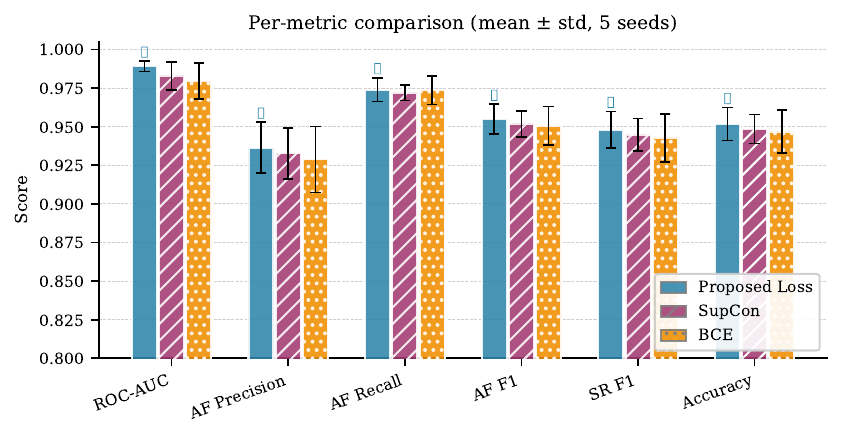}
  \caption{Per-metric linear-probe comparison of the proposed
  patient-aware objective, SupCon, and BCE on the held-out
  patient-independent test split (5~seeds). Bars show the seed
  mean and error bars show $\pm$~one standard deviation.}
  \label{fig:loss_comparison}
\end{figure}

The geometric ordering is preserved on the downstream task
(Figure~\ref{fig:loss_comparison}). 
AUROC orders the three losses identically to per-patient SR cohesion 
(Section~\ref{subsec:repr_analysis}). 
The most striking effect is on stability. 
Seed to seed AUROC variance is reduced by $2.6\times$ over SupCon and $3.4\times$ over BCE. 
A geometrically consistent embedding space is less sensitive to weight initialization, which is a desirable property for any deployment that must reproduce model behaviour. 
AUROC improvements of $+0.006$ over SupCon and $+0.009$ over BCE each exceed one standard deviation of the corresponding baseline. 
Per-class precision/recall and a comparison to prior RRI-based AF detectors are reported in Appendix~\ref{app:loss_table} (Tables~\ref{tab:results} and~\ref{tab:related_work}).
The proposed representations exceed Andersen et al.'s patient-holdout sensitivity~\cite{andersen_deep_2019} by over $11$~percentage points despite using only a frozen encoder and a logistic-regression probe.

\section{Discussion}

\textbf{Per-subject consistency, not global separability, governs
transfer.}
The BCE paradox is the clearest demonstration of our central
claim. Class separability metrics tell us how far the class means
have been pushed apart, but a linear probe trained on seen
patients can only generalize to new ones if the embedding space
provides a consistent direction along which to extrapolate. When
the per-patient SR structure is disorganized, that direction is no
longer well-defined even if the class means themselves are far
apart. The proposed objective targets this consistency
directly by forming positives only from same-patient, same-class
segments, and its $+0.050$ SR-cohesion advantage over SupCon
translates into a $2.6\times$ reduction in AUROC variance across
seeds. Geometrically, the loss maintains a per subject balance
between alignment and uniformity. Each patient's class-conditional cluster
is tightly aligned while clusters from different patients remain spread out on the hypersphere. This avoids both
SupCon's collapse into a single shared class prototype and BCE's
cross-patient inconsistency.

\textbf{Generality and clinical relevance.}
The construction does not depend on the encoder and only requires
subject identifiers and class labels in each batch. The mechanism 
applies wherever (a) class labels are available at training time 
and (b) within-subject variation is informative. Whether the same 
gains transfer to other physiological signals organized by subject 
(EEG, EMG, PPG) remains an open empirical question and a 
direction for future work. On the clinical side, the consistent
AF recall ($0.974 \pm 0.008$) across seeds matters. A missed AF
episode may delay anticoagulation and elevate stroke risk, so
seed-stable sensitivity is a deployment advantage. 
The gain is in the representation, not the classifier. 
This downstream performance is achieved with a frozen encoder 
and a logistic regression probe and exceeds Andersen et al.'s patient-holdout sensitivity~\cite{andersen_deep_2019} by over $11$~percentage points (Appendix~\ref{app:loss_table}, Table~\ref{tab:related_work}).

\textbf{Limitations.}
Our analysis is on a single dataset (IRIDIA-AF) with a
frozen-probe protocol.
Prospectively, multi-centre validation and
end-to-end fine-tuning remain open. 
Three concrete next steps follow from the geometric findings.
\begin{itemize}
  \item A per-subject decomposition of alignment and uniformity to verify the mechanism formally.
  \item Few shot adaptation that uses each patient's preserved SR structure as a personal prior.
  \item Applying the same class and subject positive construction to other physiological signals where between subject variability is the main barrier to generalization.
\end{itemize}

\section{Conclusion}

We proposed a patient-aware contrastive objective for physiological signals that are organized by subject. 
Positives are restricted to same-patient, same-class pairs, preserving each subject's own structure while still pushing the two classes apart.
On IRIDIA-AF, the construction attains the most consistent per-patient SR cohesion among the losses we compared, uncovers a BCE paradox.
Overall class separability is a misleading indicator of how well a representation transfers to unseen subjects.
Validated downstream, the construction reaches AUROC $0.989 \pm 0.003$ for PAF detection on unseen patients with $2.6\times$ lower seed variance than supervised contrastive baselines. 
The objective does not depend on the encoder and only requires subject IDs at training time. 
This suggests same-class, same-subject positive construction as a broadly useful primitive for representation learning on signals that are organized by subject.

\section*{Acknowledgements}
We thank Joshua Pranjeevan Kulasingham for his support.

\bibliography{main}
\bibliographystyle{icml2026}

\newpage
\appendix
\onecolumn

\section{Preprocessing Pipeline}
\label{app:preprocessing}

\begin{figure}[h]
  \centering
  \includegraphics[width=0.7\linewidth]{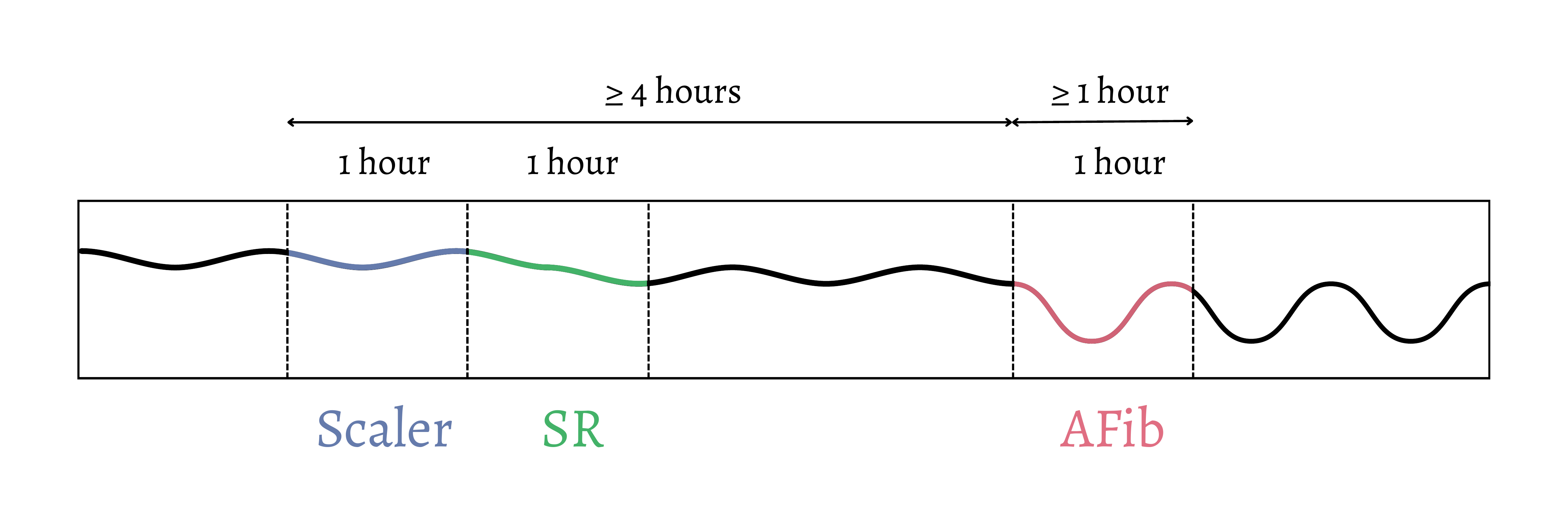}
  \caption{Episode selection and preprocessing pipeline. SR
  normalization windows (hour~0--1) are disjoint from
  classification windows (hours~1--2), ensuring the RobustScaler
  has no access to labeled data. The $\geq 4$~hr SR inclusion
  criterion further ensures that all SR classification windows
  begin at least 2~hr before AF onset, reducing the risk of
  including pre-episode transitional rhythms.}
  \label{fig:preprocessing}
\end{figure}

Per-patient normalization uses,
\begin{equation}
    \tilde{RR}_i = \frac{RR_i - \mathrm{median}(\mathcal{R}^{(p)}_{\mathrm{scale}})}{\mathrm{IQR}(\mathcal{R}^{(p)}_{\mathrm{scale}})}
    \label{eq:norm}
  \end{equation}
where $\mathcal{R}^{(p)}_{\mathrm{scale}}$ is the patient specific
fit window (the first SR hour). Beats with RR $< 200$\,ms or
$> 2000$\,ms are excluded as physiologically implausible. The
hyperparameter optimization is described in Appendix~\ref{app:training}.

\section{Encoder Architecture and Training Details}
\label{app:training}

\begin{figure}[h]
  \centering
  \includegraphics[width=0.7\linewidth]{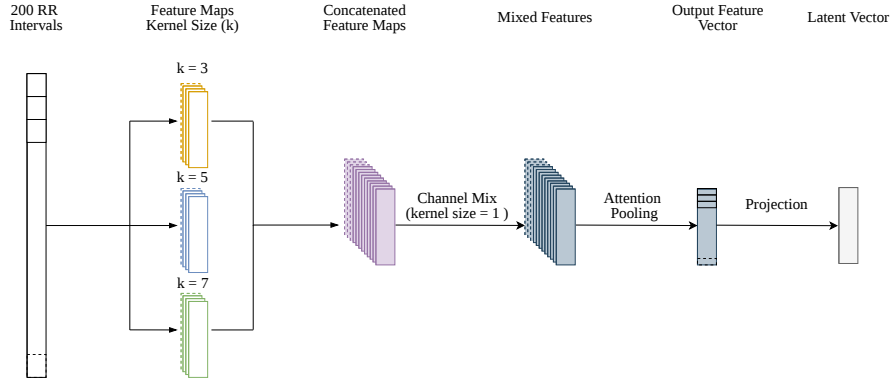}
  \caption{Encoder architecture: multi-branch 1D-CNN backbone with
  temporal attention pooling and a two-layer MLP projection head.}
  \label{fig:encoder}
\end{figure}

\textbf{Architecture.} Three parallel 1D-CNN branches with kernel
sizes $k\!\in\!\{3,5,7\}$ capture beat-to-beat fluctuations,
medium-range oscillations, and broader trend dynamics, motivated
by Inception style multi-scale processing~\cite{szegedy_going_2015}. 
Each branch applies three strided convolution blocks (stride 2) with channel depths
$\{16,32,64\}$, Group Normalization~\cite{wu_group_2018} (8
groups), and ReLU. Branch outputs are concatenated and fused via a
$1{\times}1$ convolution, then pooled by a softmax temporal
attention module~\cite{shashikumar_detection_2018},
\begin{equation}
\alpha_t = \frac{\exp(\mathbf{w}^\top \mathbf{h}_t)}{\sum_{t'} \exp(\mathbf{w}^\top \mathbf{h}_{t'})}, \quad
\mathbf{z} = \sum_t \alpha_t \mathbf{h}_t.
\end{equation}
A two-layer MLP projects $\mathbf{z}\in\mathbb{R}^{128}$ to
$\mathbf{e}\in\mathbb{R}^{128}$; both are $\ell_2$-normalized to
the unit hypersphere~\cite{wang_understanding_2020}. The
contrastive loss operates on $\hat{\mathbf{e}}$; the linear probe
on $\hat{\mathbf{z}}$.

\textbf{Training.} PyTorch on a single NVIDIA RTX 2080. AdamW with
learning rate $6.8{\times}10^{-3}$ and weight decay
$8.8{\times}10^{-4}$; cosine annealing to $10^{-6}$; dropout
$0.12$ on the projection head; learnable temperature $\tau$
initialized at $0.05$. Patient-aware sampling uses $P=8$ patients
and $n=16$ windows per class per-patient ($B=128$). Up to 100
epochs with early stopping on validation AUROC (patience 10).

\textbf{Hyperparameter selection.} 32 hyperparameters
(architecture, optimizer, data pipeline including $W$ and $S$)
are jointly tuned by Optuna's TPE
sampler~\cite{akiba_optuna_2019} on the validation split, with
the test split held out until final evaluation. For all three loss conditions (Proposed, SupCon, BCE), only the 
temperature~$\tau$ and learning rate are varied to find the best 
performance per loss. The encoder architecture, sampler, and probe 
are held fixed across all conditions, giving no tuning advantage 
to the proposed loss.

\section{Embedding-Geometry Metric Definitions}
\label{app:metric_defs}

The encoder produces L2-normalized embeddings, so every embedding
$z_i\!\in\!\mathbb{S}^{d-1}$ satisfies $\|z_i\|\!=\!1$. Let
$y_i\!\in\!\{\text{SR},\text{AF}\}$ denote the class label and
$p_i\!\in\!P$ the patient identifier. Define the index sets
\[
S^c \;=\; \{i : y_i = c\},
\qquad
S_p^c \;=\; \{i : y_i = c,\, p_i = p\},
\]
the (unnormalized) class centroids
\[
\mu^c \;=\; \frac{1}{|S^c|}\sum_{i\in S^c} z_i,
\qquad
\mu_p^c \;=\; \frac{1}{|S_p^c|}\sum_{i\in S_p^c} z_i,
\]
their L2-normalized versions
$\hat\mu^c\!=\!\mu^c/\|\mu^c\|$,
$\hat\mu_p^c\!=\!\mu_p^c/\|\mu_p^c\|$, and the mean class spreads
\[
\bar d^c \;=\; \frac{1}{|S^c|}\!\sum_{i\in S^c}\!\|z_i-\mu^c\|_2,
\qquad
\bar d_p^c \;=\; \frac{1}{|S_p^c|}\!\sum_{i\in S_p^c}\!\|z_i-\mu_p^c\|_2.
\]

\paragraph{Per-patient class cohesion (primary metric).}
For class $c\!\in\!\{\text{SR},\text{AF}\}$,
\begin{equation}
\mathrm{Coh}^{\,c}
\;=\;
\frac{1}{|P|}\sum_{p\in P}\;
\frac{1}{|S_p^c|}\sum_{i\in S_p^c} z_i^{\!\top}\hat\mu_p^{\,c}.
\label{eq:cohesion}
\end{equation}
Because $\|z_i\|\!=\!\|\hat\mu_p^c\|\!=\!1$, each summand is the
cosine similarity between an embedding and its same-patient
same-class centroid; values lie in $[-1, 1]$ and approach $1$ as
the per-patient class cluster tightens.

\paragraph{Global class separability.}
\begin{align}
\mathrm{CentDist} &= \|\mu^{\text{SR}} - \mu^{\text{AF}}\|_2,
\label{eq:centdist}\\[2pt]
\mathrm{CentSim}  &= \hat\mu^{\text{SR}\,\top}\hat\mu^{\text{AF}},
\label{eq:centsim}\\[2pt]
C_{\text{glob}}   &= \frac{\|\mu^{\text{SR}} - \mu^{\text{AF}}\|_2}
                          {\bar d^{\text{SR}} + \bar d^{\text{AF}}}.
\label{eq:cglobal}
\end{align}
We refer to $C_{\text{glob}}$ as a \emph{compactness ratio}. It is
the ratio of between-class centroid distance to the sum of
within-class mean spreads. 

\paragraph{Per-patient compactness.}
\begin{equation}
C_{\text{pp}}
\;=\;
\frac{1}{|P|}\sum_{p\in P}
\frac{\|\mu_p^{\text{SR}} - \mu_p^{\text{AF}}\|_2}
     {\bar d_p^{\text{SR}} + \bar d_p^{\text{AF}}}.
\label{eq:cpp}
\end{equation}
$C_{\text{pp}}$ measures the same separation-to-spread trade-off
as $C_{\text{glob}}$, but evaluated within each patient and then
averaged.

\emph{Note:} when within-patient class spread is very small
($\bar d_p^{\mathrm{SR}} + \bar d_p^{\mathrm{AF}} \to 0$),
the ratio becomes large regardless of whether the resulting clusters are
positioned consistently across patients. In such cases $C_{pp}$ should be
interpreted alongside the raw spreads $\bar d_p^c$ and the per-patient
cohesion scores rather than in isolation.

\section{Embedding-Geometry Metrics (Full Table)}
\label{app:repr_metrics}

\begin{table*}[h]
  \centering
  \caption{Embedding-space metrics on the test set
  (mean $\pm$ std, 5 seeds). Per patient SR cohesion is the
  primary metric. Bold = best per row.
  $\uparrow$ higher is better; $\downarrow$ lower is better.}
  \label{tab:appendix_repr_metrics}
  \begin{small}
    \begin{tabular}{lccc}
      \toprule
      \textbf{Metric}
        & \textbf{Proposed}
        & \textbf{SupCon}
        & \textbf{BCE} \\
      \midrule
      SR cohesion $\uparrow$
        & $\mathbf{0.850 \pm 0.044}$
        & $0.800 \pm 0.029$
        & $0.772 \pm 0.049$ \\
      AF cohesion $\uparrow$
        & $0.846 \pm 0.048$
        & $0.921 \pm 0.035$
        & $\mathbf{0.955 \pm 0.019}$ \\
      Centroid similarity $\downarrow$
        & $+0.238 \pm 0.200$
        & $+0.155 \pm 0.123$
        & $\mathbf{-0.717 \pm 0.029}$ \\
      Global centroid dist. $\uparrow$
        & $1.034 \pm 0.100$
        & $1.120 \pm 0.095$
        & $\mathbf{1.603 \pm 0.045}$ \\
      Global compactness ratio $\uparrow$
        & $1.192 \pm 0.124$
        & $1.313 \pm 0.465$
        & $\mathbf{2.427 \pm 0.226}$ \\
      Per-patient compactness ratio $\uparrow$
        & $\mathbf{3.273 \pm 0.433}$
        & $2.178 \pm 1.148$
        & $6.396 \pm 0.590$ \\
      \midrule
      \textbf{AUROC} $\uparrow$
        & $\mathbf{0.989 \pm 0.003}$
        & $0.983 \pm 0.009$
        & $0.980 \pm 0.012$ \\
      \bottomrule
    \end{tabular}
  \end{small}
\end{table*}

\section{Downstream PAF Detection (Per-Class Metrics and Loss Comparison)}
\label{app:loss_table}

\begin{table}[h]
  \centering
  \caption{Per-class linear-probe metrics for the proposed
    representations (4160~AF and 3799~SR segments from unseen
    patients; mean~$\pm$~std across 5 seeds). The headline AUROC
    of $\mathbf{0.989\pm0.003}$ is reported in the main text.}
  \label{tab:results}
  \begin{small}
    \begin{tabular}{lccc}
      \toprule
      \textbf{Class} & \textbf{Prec.} & \textbf{Recall} & \textbf{F1} \\
      \midrule
      SR (0)
        & $0.970 \pm 0.009$
        & $0.927 \pm 0.020$
        & $0.948 \pm 0.012$ \\
      AF (1)
        & $0.936 \pm 0.017$
        & $\mathbf{0.974 \pm 0.008}$
        & $0.955 \pm 0.010$ \\
      \midrule
      \multicolumn{2}{l}{Accuracy}
        & \multicolumn{2}{c}{$0.952 \pm 0.011$} \\
      \multicolumn{2}{l}{\textbf{AUROC}}
        & \multicolumn{2}{c}{$\mathbf{0.989 \pm 0.003}$} \\
      \bottomrule
    \end{tabular}
  \end{small}
\end{table}

\begin{table}[h]
  \centering
  \caption{Linear-probe comparison across loss functions on the
  fixed patient-independent test split (5~seeds; bold = best mean
  per metric).}
  \label{tab:loss_comparison}
  \setlength{\tabcolsep}{4pt}
  \small
  \begin{tabular}{lcccccc}
    \toprule
    \textbf{Loss}
      & \textbf{AUROC}
      & \textbf{AF Prec.}
      & \textbf{AF Recall}
      & \textbf{AF F1}
      & \textbf{SR F1}
      & \textbf{Acc.} \\
    \midrule
    Proposed
      & $\mathbf{0.989\!\pm\!0.003}$
      & $\mathbf{0.936\!\pm\!0.017}$
      & $0.974\!\pm\!0.008$
      & $\mathbf{0.955\!\pm\!0.010}$
      & $\mathbf{0.948\!\pm\!0.012}$
      & $\mathbf{0.952\!\pm\!0.011}$ \\
    SupCon
      & $0.983\!\pm\!0.009$
      & $0.933\!\pm\!0.017$
      & $0.972\!\pm\!0.005$
      & $0.952\!\pm\!0.008$
      & $0.945\!\pm\!0.011$
      & $0.949\!\pm\!0.009$ \\
    BCE
      & $0.980\!\pm\!0.012$
      & $0.929\!\pm\!0.021$
      & $\mathbf{0.974\!\pm\!0.009}$
      & $0.951\!\pm\!0.012$
      & $0.943\!\pm\!0.016$
      & $0.947\!\pm\!0.014$ \\
    \bottomrule
  \end{tabular}
\end{table}


\section{Comparison with Prior Work}

\begin{table}[h]
  \centering
  \caption{Performance comparison with related work on RR-interval
  AF detection. Direct numerical comparison is limited by dataset
  and evaluation differences.
  $^\dagger$ Cross-validation with within-patient data mixing
  inflates reported metrics.
  $^\ddagger$ MIT-BIH~AF contains no episodes satisfying our
  quality criteria (AF~$\geq\!1$h, preceding SR~$\geq\!4$h).}
  \label{tab:related_work}
  \setlength{\tabcolsep}{3pt}
  \small
  \begin{tabular}{llllccc}
    \toprule
    \textbf{Study} & \textbf{Method} & \textbf{Dataset}
      & \textbf{Validation} & \textbf{Sens.}
      & \textbf{Spec.} & \textbf{Acc.} \\
    \midrule
    \cite{udawat_automated_2022}
      & HRV+ML & MIT-BIH~AF
      & CV$^\dagger$ & 95.16\% & 92.46\% & 94.43\% \\
    \cite{andersen_deep_2019}
      & CNN+RNN & 3 dbs
      & 5-fold CV$^\dagger$ & 98.98\% & 96.95\% & --- \\
    \cite{andersen_deep_2019}
      & CNN+RNN & Unseen
      & Pt.\ holdout & 86.04\% & 98.96\% & --- \\
    \cite{faust_automated_2018}
      & LSTM & MIT-BIH$^\ddagger$
      & 10-fold CV$^\dagger$ & --- & --- & 98.51\% \\
    \midrule
    \textbf{Ours}
      & CNN+MLP & IRIDIA-AF
      & Pt.\ holdout
      & $\mathbf{97.40\!\pm\!0.80\%}$
      & $\mathbf{92.72\!\pm\!2.01\%}$
      & $\mathbf{95.17\!\pm\!1.07\%}$ \\
    \bottomrule
  \end{tabular}
\end{table}

\end{document}